# A Fuzzy Realistic Mobility Model for VANET

**Alireza Amirshahi[1], Mahmood Fathi[2], Morteza Romoozi[3] and Mohammad Assarian[4]**

[1] **Computer Engineering Department, Islamic Azad University, Arak Branch**
**Arak, Iran**

[2] **Computer Engineering Department, Science and Technology University, Tehran Branch**
**Tehran, Iran**

[3,4] **Computer Engineering Department, Islamic Azad University, Kashan Branch**
**Kashan, Iran**

**Abstract**

Realistic mobility models can assess more the results more accurate estimate parameters because it is closer to the real world. In this paper a realistic Fuzzy Mobility Model has been proposed. This model has rules which are changeable depending on nodes and environmental conditions. This model is more complete and precise than the other mobility models. After simulation, it was found out that not only considering nodes movement as being imprecise (fuzzy) has a positive effects on most of ad hoc network parameters, but also, more importantly as they are closer to the real world condition, they can have a more positive effect on the implementation of ad hoc network protocols.

## 1. Introduction

Nowadays ad hoc networks have been used in a variety of applications. Mobility models in ad hoc networks are of special importance. Mobility model identifies the primary place of nodes and the manner of nodes mobility. Mobility models fall into two categories: realistic and unrealistic. As realistic mobility models are more similar to real world conditions, they provide more accurate results.

Vehicular ad-hoc networks (VANETs) are a particular kind of mobile ad-hoc networks where nodes are embedded into Moving vehicles, equipped with short-range wireless communication devices and positioning systems like GPS or Galileo. VANETs are gaining popularity in both academia and Industry as a key technology for many emerging services and applications in the automotive field, e.g. safety, traffic optimization and infotainment.

Before applying to the real world, computer simulation is a valuable tool for evaluating protocols and other network parameters. Simulation can be applied easily, while implementation of ad hoc networks in the real world is difficult and expensive. Moreover, simulation has other advantages such as iterative scenario, parameter isolation, and measuring different metrics. Glomosim [1] and NS2 [2] are the most famous simulators used for evaluating and comparing computer network protocols. Mobility model, signal propagation model and routing protocol are the most important parts of wireless simulators. Many realistic models have been presented in most of which nodes mobility is random and simulation environment is free, without obstacle and pathway.

Applying these models cannot represent the efficiency of the networks in real condition because in real condition, nodes must move in predefined passages and nodes signals must be blocked by obstacles. The movement patterns and path selections are not random. Some realistic models have been presented so far like Graph-based Mobility Model [3] and Obstacle Model [4] and. In these models, there are usually obstacles and pathway, but no attention has been paid to the movement patterns and destination selection. Meanwhile, the type of destination selection of nodes is not random. For instance, the selection of people’s destination in a VANET environment is not random and many parameters are involved such as time, current place, the priority of going to different places and etc.

The mobility of a mobile node and its mobility environment are not precise. Namely, a urban environment is not precise because every place has different parts and precise coordination of each part cannot be stated.

As fuzzy control systems are capable of solving imprecise problems efficiently, by using fuzzy control system in the proposed Fuzzy Mobility Model, the motion rules of different kinds of nodes, based on type of the activity and environment, have been designed. Fuzzy control system includes fuzzy rules which describe the nodes mobility in an adaptable way with the environment. This model has a knowledge base which can be changed based on nodes conditions, types of nodes and environment. By using such knowledge base, the mobility rules of every environment can be imposed upon a mobility model as an input, until the mobility is created in that specific environment.

A review of the related studies has been presented in part 2. Part 3 contains the proposed Fuzzy Mobility Model. The simulator (Glomosim[5]) and its results have been presented in part 4 and the conclusion has been mentioned in part 5.

## 2. A Review of The Related Studies

Regarding realistic mobility models, many studies have been done, but most of them have been performed on environment model and signal blockage and just a few attention has been paid to real movement patterns. In these models, destination selection of nodes was either completely random and the selection of path by algorithm was either the shortest one or it was selected randomly which are not considered suitable. For instance, the Obstacle Mobility Model [4], presented in 2003 by A. Jardesh, is one of the most successful realistic models. This model has an appropriate signal blockage and environment sub-model and it can be used as signal blockage and environment sub-model of other models, but it lacks a real world-based mobility pattern model. The destination selection of this model is entirely random and the path selection is done by Dijkstra algorithm with the shortest path regarding the number of edges which is not a suitable criterion.

A realistic group model, called OCGM [6], based on obstacle and mobility model RPGM [7] has been proposed which has similar environment model and blockage signal, but in its movement pattern sub-model, nodes move in groups.

Graph-based [3] is another model, environment model of which is constituted by a graph and this graph is the paths of a map and has not a specific signal blockage sub-model.

The next model which is based on Graph-based Model, named Area Graph-based [8], has been represented and its environment sub-model is similar to Graph-base Model and lacks signal blockage model, but compared to Graph-based Model, its movement pattern sub-model has been improved. In the other words as long as the nodes are inside the graph vertices, they have Random Waypoint [9] mobility. But for leaving vertex, nodes must select one of the output edges of the vertex which has probability from the beginning of simulation, along with related probability.

Still another realistic model, called Environment Aware Mobility [10] has been represented. The environment sub-model of this model is different from that of Graph-based Model mentioned above. In this mobility model, the environment is divided into a series of sub-environment inside of which there are some obstacles and movement pattern sub-model of nodes in each sub-environment can be one of the random mobility models. This model has signal blockage sub-model.

There are some realistic mobility models which touch on nodes movement pattern models. But the number of these models is by far fewer than the other models. For example, we can refer to a Cluster-based Mobility Model [11] for intelligent nodes by M.Romoozi. This mobility model has focused on the movement pattern sub-model and has improved it.

H. Babaei has proposed another model in 2007, named Obstacle Mobility Model Based on Activity Area [12]. This model has used environment sub-model and signal obstruction model of obstacle, but in this model the node movement pattern has been improved and a different range of activity and speed has been assigned to each group of nodes. Nodes select those vertices which are closer to the area of the activity with greater probability, but this selection is done by Dijkstra algorithm.

There are other models such as Manhattan Mobility Model [13], Free Way Mobility [13], and Urban Mobility Model [14]. But compared to the more complete models mentioned above, these models are of little importance.

### 2.1 Classification of Mobility Models

Mobility models can be divided into two categories: realistic and unrealistic. In realistic mobility models, the mobility of nodes is assessed in the real world conditions. In this model, not only mobility pattern of mobile nodes is considered, but also simulation environment and the effect of environment on signal nodes are examined. In unrealistic mobility models, a free and without obstacle space is taken into account in which nodes move freely everywhere and their selection of destination and path is usually random and there is no predefined obstacle and pathway for them. These models do not determine an accurate result in evaluating ad hoc network protocols.

#### 2.1.1 Realistic Mobility Models

Realistic mobility models create an environment similar to the real world. This environment includes some pathways through which nodes must move in these pathways. This model also includes some obstacles. Not only these obstacles obstruct the nodes movement, but also they weaken or remove nodes signal. The more the environment is similar to the real world conditions, the more accurate the evaluation results will be.

Considering the analyses have done up to now, each realistic model is usually composed of 3 sub-models. These sub-models are related to one another and they can hardly be separated as follow:

- -- Environment sub-model.
- -- Signal obstruction sub-model.
- -- Movement pattern sub-model.

Environment sub-model includes environmental obstacles such as buildings, mountains and etc., which usually exist in real environment. This sub-model also includes some paths that exist among these obstacles. These paths force the nodes to move only through these paths.
Signal obstruction sub-model in the proposed mobility model does not include obstacles.
Movement pattern sub-model includes the manner of destination selection, the selection of path toward the destination, and the amount of pause in destination. To sum up, the manner nodes movement is explained in environment sub-model.

### 2.2 Fuzzy Control System

The term ‘fuzzy’ means imprecise. Although fuzzy systems describe uncertain and unclear phenomena, Fuzzy theory is a precise one. The heart of a fuzzy system is a knowledge base which is composed of fuzzy If-Then rules.
The main structure of fuzzy systems is shown in figure 1.

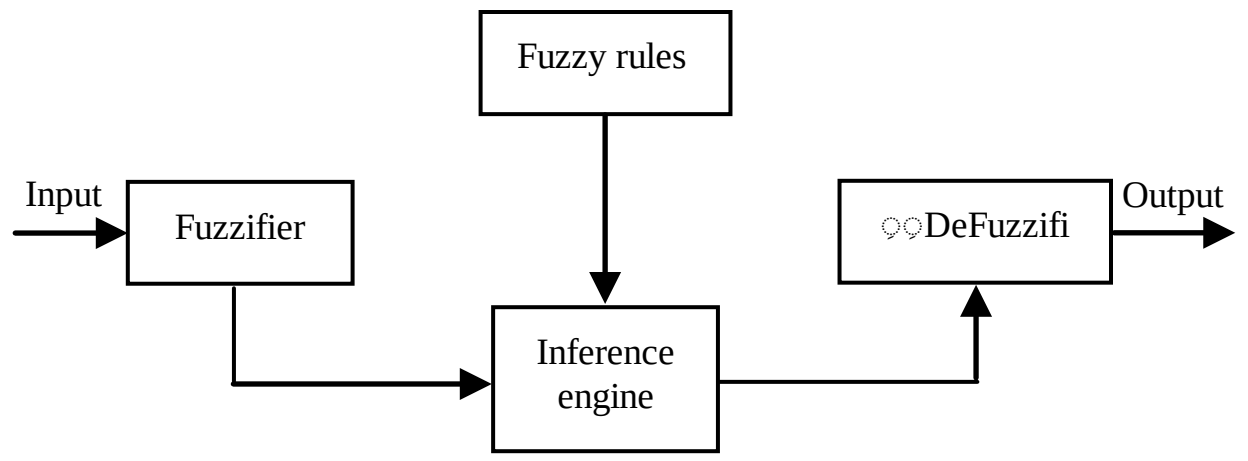

Fig. 1 The main structure of fuzzy systems.

The fuzzifier used in Fuzzy Mobility Model is a unique fuzzifier (1). This fuzzifier maps a singular point $x^* \in u$ with real value on a fuzzy unique A′ in u and the membership function in X* equals one and in other points u equals 0. It means:

$$u_{A'}(x) = \begin{cases} 1 : x = x^* \\ 0 : \text{O.W} \end{cases} \quad (1)$$

Defuzzifier used in Fuzzy Mobility Model is the center average defuzzifier. Center average defuzzifier is the most commonly used defuzzifier in fuzzy systems and fuzzy control systems. Fiscally it is simple, and at the same time, intuitively it is justifiable. Center average defuzzifier can be defined in the following way:

$$y^* = \frac{\sum_{\ell=1}^{M} y^{-\ell} w_\ell}{\sum_{\ell=1}^{M} w_\ell} \quad (2)$$

In this equation (2) $y^{-\ell}$ is the center of $\ell$ fuzzy set and $w_\ell$ is its hight degree and M is the number of our rules.
Inference engine of used in the fuzzy mobility model is multiple inference engine (3).

$$u_{B'}(y) = \max_{\ell=1}^{M} \left[ \sup_{x \in u} \left( u_{A'}(x) \prod_{i=1}^{n} u_{A_i}\ell(x_i) u_B \ell(y) \right) \right] \quad (3)$$

## 3. The Proposed Fuzzy Mobility Model

In a real environment, nodes are divided into different groups with similar mobility features. For instance, in VANET we have personal automobile nodes, public automobile nodes, and ambulance automobile nodes. For each group of nodes, the manner of destination selection, the movement speed, time and etc., are different.
In the real world, the destination of nodes is expressed imprecisely and the conception of fuzziness is hidden in it. For example, emergence place is included different sections, and we cannot express a precise coordinates. For example, the emergence place cannot be stated in unique X and Y points.
In the real world, nodes destination is selected based on the time. Namely, in VANET, a personal automobile goes to a university in the morning and to the residential complex at noon, but these times are not stated exactly. Some people believe that morning starts from 7 to 10, but others believe it to be from 7 to 9 and etc. So time can be considered as being fuzzy.
Regarding that each of the nodes has a different mobility, so the destination selection of each node will be different from the others. To provide an example, in a VANET environment, the mobility of a personal automobile node is different from that of a public or an ambulance. Therefore, their destinations are different.

### 3.1 Sub-Models of Fuzzy Mobility Model

Sub-models of Fuzzy Mobility Model include 3 parts:

#### 3.1.1 Environment Sub-Model

In environment sub-model of the proposed mobility model is used a city map that Pathways in this sub-model are like a real environment. In fact we create a real environment to have a real simulated environment. In figure 2, buildings have been created like Squares and available edges in that are some pathways.

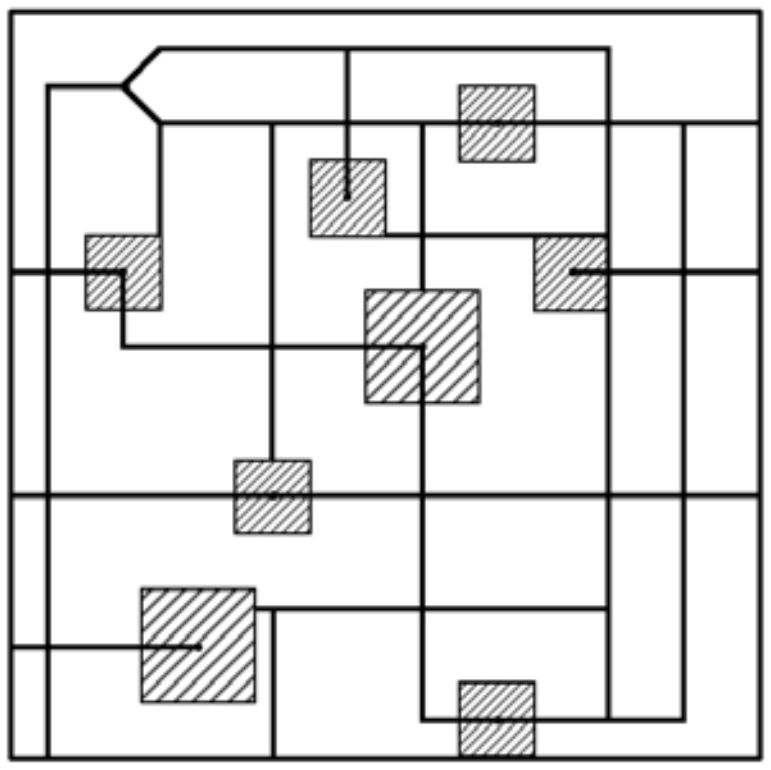

Fig. 2 Simulation environment.

### 3.1.2 Signal Obstruction Sub-Model

Whereas of the mobility model proposed there are no obstacles, in this mobility model, Signal obstruction sub-model does not use.

### 3.1.3 Movement Pattern Sub-model

The manner of movement, including the path selection, destination selection, and the amount of pause in destination, will be examined in this sub-model.

In this paper, the main focus is on the movement pattern of nodes. In the proposed Fuzzy Mobility Model, firstly, nodes are divided into groups with equal mobility features. Then, the manner of destination selection of nodes is defined by using a fuzzy control system (fuzzifier, fuzzy rules, inference engine and defuzzifier).This mobility model is suitable for mobility of a mobile node which has an imprecise mobility. The pass selection method in proposed mobility model is Dijkstra shortest pass algorithm.

The proposed mobility model gains from a fuzzy control system that contains fuzzy rules. These fuzzy rules describe the node mobility in an adaptable way to the environment. Fuzzy rules express the manner of destination selection for each group of nodes. As time and place inputs are expressed imprecisely (fuzzy), it seems that Fuzzy Mobility Model is more similar to the real world than the previous realistic model and this model can be used as a part of a simulator by MANET network researchers.

In the proposed Fuzzy Mobility Model, considering fuzzy systems (figure.1), fuzzifier input is time and place which are expressed precisely (for example for ambulance automobile node, the current place is hospital and the time is 8 o'clock). Fuzifier changes these amounts from being precise into fuzzy state and they go to the inference engine along with current fuzzy rules. Afterwards, the output of the inference engine goes to the defuzzifier and the next destination is defined based on the fuzzy rules.

## 3.2 Nodes Clustering

In VENET environment nodes can be divided into 3 groups: personal, public and ambulance. Each of these nodes has different motilities which are explained later on.

## 3.3 Mobility Analysis

Mobility analysis has different methods including locating the camera in specific places and identifying the movement of people and obtaining the related mobility model and the other method is using Radio-Frequency Identification (RFID). Still the next method is using questionnaire which has been in this mobility model. We distributed questionnaires among some VANET nodes and asked them to fill out the forms. This way we were able to identify the destination of these nodes in different time intervals. Because each node has different programs, the fuzziness of mobility model is proved.

## 3.4 Inputs of Fuzzy Mobility Model (VANET environment)

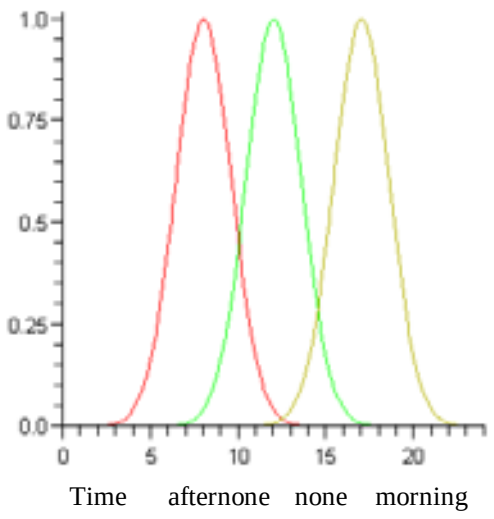

Fig. 3 Time input in Fuzzy Mobility Model.

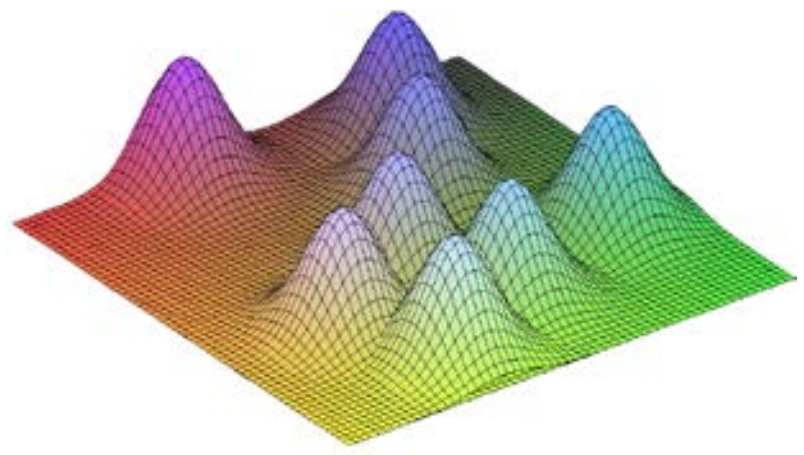

Fig. 4 Places of Fuzzy Mobility Model.

### 3.4.1 Time Input

Time input is divided into 3 parts: morning, noon and evening (according to figure. 3 mapped by Maple Software). A Gausian diagram has been applied to show the time as being fuzzy and its membership function has been shown in (4). A is a point in which the diagram has the highest value '1'. For instance, the value of a in the morning, at noon, and in the evening can be 8, 12, and 17 respectively.

$$\mu_{time}^{(t)} = e^{(-0.2(x-a)^2)} \quad (4)$$

3.4.2 Place Input

The other input is place which is mapped by maple software (figure. 4). In order to show the place as being fuzzy, a Gausian diagram has been used and its membership function has been shown in (5). a and b are the coordinates of the center of places and their diagram in that points has the highest value. Coordinates of the center of sites has been shown in figure. 5.

$$\mu_{pos}^{(x,y)} = e^{(-(10^{-4}((x-a)^2+(y-b)^2)))} \quad (5)$$

## 3.5 The output of the Fuzzy Mobility Model

The output of the Fuzzy Mobility Model is the selection of destination. Considering figure. 1(the main structure of Fuzzy Systems), the inputs of mobility model are the current time and place given to the fuzzy control system (given precisely). Now regarding the given rules table, the place of destination is defined.

## 3.6 Extracting the Rules Table

As mentioned before, VANET nodes have different mobilities in different times. So questionnaires have been used. Table 1 shows a type of the forms used for each of the nodes.

Table 1: Questionnaire forms of VANET nodes.

| Question Form | Public automobiles ☐ | personal automobiles ☐ | Ambulance automobiles ☐ |
|---|---|---|---|

| Morning | | Noon | | Evening | |
|---|---|---|---|---|---|
| Place | Priority | Place | Priority | Place | Priority |
| Hospital | | Hospital | | Hospital | |
| Emergency | | Emergency | | Emergency | |
| University | | University | | University | |
| Residential | | Residential | | Residential | |
| -Complex | | -Complex | | -Complex | |
| Park | | Park | | Park | |
| Bazaar | | Bazaar | | Bazaar | |
| City center 1 | | City center 1 | | City center 1 | |
| City center 2 | | City center 2 | | City center 2 | |

Questionnaire forms were submitted to 70 personal automobiles, 70 public automobile and 70 ambulance automobile and they were all asked to read and fill out the forms carefully. For example, the ambulance automobile has to define the priority of going to the class of hospital by a digit between 0 and 1 for the morning, noon, and the evening in the questionnaire. He was asked to do the same for other places in the form. When the nodes returned the questionnaire, the average of priority of each node in the morning, at noon, and in the evening was calculated and this knowledge was used for problem solving.

In order to fill out the rules table, a map of the real urban environment is provided. Then, the centers of sites are defined by exact X and Y in a coordinate axis in which X and y have the maximum value of 10000. In figure. 5, the precise center of VANET sites is shown. For instance emergence place is located in coordinates X= 7500 and Y=6500.

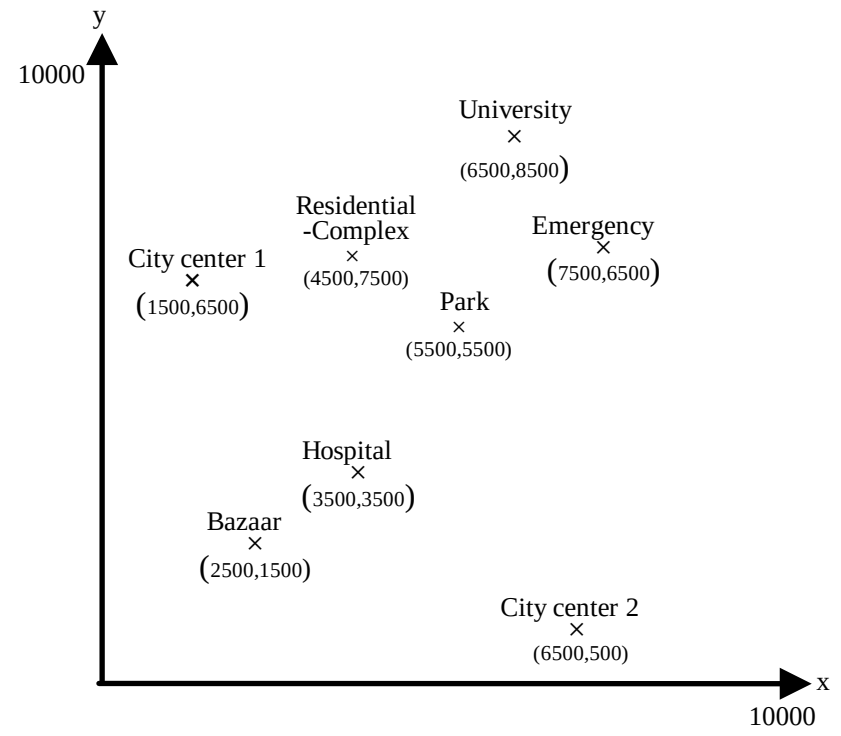

Fig. 5 Places coordinate of Fuzzy Mobility Model in VANET environment.

There are two important parameters in the proposed mobility model. These parameters are required for deciding the direction of nodes movement from one place to the other such as priority of nodes for moving towards the destination sites and distance of nodes from destination sites. In (6), both parameters have been taken into account. $P_1$ and $P_2$ are used for defining the weight of these parameters. This is a minimum equation.

$$\underset{Site=1}{\overset{n}{Min}}\, k = P_1(1-A) + P_2\left(\frac{d}{Max_dist}\right) \quad (6)$$

In this equation A is the priority of going to a place extracted from the questionnaire and d is the distance between the current place and node destination. As we have the coordinates of the center of sites, the distance between these two sites can be obtained from equation $d = \sqrt{(x_2-x_1)^2+(y_2-y_1)^2}$ ($x_1$ and $y_1$) are the coordinates related to the current site and ($x_2$ and $y_2$) are the coordinates of centers of the main destinations.(Current place and destination place are the sites shown in figure. 5). The distance between two sites (d) is divided by the maximum distance (Max_dis). According to this plan the farthest distance between two places equals 8060 (Max_dis=8060), so the results will be a digit between 0 and 1 (6). In conclusion, the more the result d/Max_dis is, the more the distance between two sites will be.

In (6), $P_1$ and $P_2$ are respectively the priority of going to the destination place and giving priority to the current place rather than destination place "$0 \le p_1, p_2 \le 1$". As the priority of going to a place is more important, the value of $P_1$ is considered equal to 0.6. Regarding the result of d/Max_dis, instead of using A, (1-A) is used to create a balance in the equation. Now, the more the result of the equation $p_1$(1-A)+$P_2$(d/Max_dis), it points to the fact that the selection of this destination is not an appropriate choice .In each place we are, this equation must be repeated for each of other places (that can be one of the destination places) and finally the obtained figure, which has the least value, is selected as destination.

### 3.7 The Calculations Done in the VANET Environment

As we are in the current place and because of having 8 existing places figure. 5 in the current place, for destination selection, we should apply (6) 8 times and the selected destination will be the result in these 8 steps. Namely, in table 1 the node is personal automobile, so if the current place is the hospital and the time is morning, the destination place will be city center 2.
Tables 2, 3, and 4 show the rules of the personal, public, and ambulance nodes in Fuzzy Mobility Model.

Table 2: Personal automobile

| place \ Time | Morning | Noon | Evening |
|---|---|---|---|
| Hospital | City center 2 | City center 1 | Bazaar |
| Emergency | City center 2 | Residential-Complex | Park |
| University | University | Residential-Complex | Residential-Complex |
| Residential-Complex | City center 1 | Residential-Complex | Park |
| Park | City center 1 | City center 1 | Park |
| Bazaar | Bazaar | City center 2 | Bazaar |
| City center 1 | City center 1 | City center 1 | City center 1 |
| City center 2 | City center 2 | City center 2 | City center 2 |

Table 3: Public automobile

| place \ Time | Morning | Noon | Evening |
|---|---|---|---|
| Hospital | City center 1 | Residential-Complex | Bazaar |
| Emergency | Residential-Complex | Emergency | Emergency |
| University | Residential-Complex | Residential-Complex | University |
| Residential-Complex | Residential-Complex | Residential-Complex | City center 1 |
| Park | City center 1 | Residential-Complex | City center 1 |
| Bazaar | Bazaar | Bazaar | City center 2 |
| City center 1 | City center 1 | City center 1 | City center 1 |
| City center 2 | City center 2 | City center 2 | City center 2 |

Table 4: ambulance automobile

| Place \ Time | Morning | Noon | Evening |
|---|---|---|---|
| Hospital | Hospital | Hospital | Hospital |
| Emergency | Emergency | Emergency | Emergency |
| University | Emergency | Emergency | Emergency |
| Residential-Complex | Emergency | Residential-Complex | Emergency |
| Park | Hospital | Emergency | Emergency |
| Bazaar | Hospital | Hospital | Hospital |
| City center 1 | Hospital | City center 1 | Hospital |
| City center 2 | City center 2 | City center 2 | City center 2 |

It should be mentioned that obtaining rules table has different ways. We can seek help from experts, for example, to complete the rules table.

## 4. Simulation

The applied simulator is called Glomosim [5].

### 4.1 Simulation Parameters

The simulation environment is 10000 m × 10000 m and the least range for transfer of nodes is 250 m. Of course, because of the existence of the obstacles, the real transmission range of each node is limited. The propagation model is two-ray path loss. In MAC layers, IEEE 802.11 DCF protocol is applied and the band is 2mbps wide. As the nodes can be pedestrian and automobile we select the mobility speed of nodes between 0 m/s and 10 m/s. The stopping time will be selected randomly between 10 and 300 seconds. In different primary situations, each point of the diagram obtained from the average 20 time-simulation implementation with distributed nodes.
After the primary distribution of nodes in the vertices of Voronoi graph, nodes move for 60 seconds to be distributed all over the simulation environment. Then, 20 Data Session begins. The size of the data packet is 512 byte and the rate of transfer is 4 packets per second. The maximum number of packet which can be sent in each data session is 6000. So a heap of 6000 packet can be received by 20 destinations. Twenty sources and destinations are selected randomly. During the simulation, the movement continues for a period of 3600 second. All the data sessions apply CBR traffic model (a fixed bit rate). The numbers of clients and servers have been selected randomly.

### 4.2 The Manner of Fuzzy Mobility Model Application to Glomosim

The formula of the fuzzy systems [15] with inference engine of multiplication, unique fuzzifier (1) and center average defuzzifier (2) will be as follow (7):

$$f(x) = \frac{\sum_{l=1}^{M} \bar{y}^l \left(\prod_{i=1}^{n} \mu_{A_i^l}(x_i)\right)}{\sum_{l=1}^{M} \left(\prod_{i=1}^{n} \mu_{A_i^l}(x_i)\right)} \quad (7)$$

$x \in U \subset R^n$ Is the input of fuzzy system and $f(x) \in V \subset R$ is the output of fuzzy system. In Fuzzy Mobility Model the above mentioned formula is implemented in C language and then it is given to Glomosim simulator.

## 4.3 Evaluation Metrics

The main purpose of simulation is the examination and comparison of evaluation metrics. Fuzzy Mobility Model in the VANET environment is compared to other mobility models. Simulation has been done according to different speeds and now we examine the results. The evaluation metrics in the simulation done are as follow:

--Node Density: The average number of each node’s neighbors is called node density.

--Broken Link Average: It is the average of broken links during the simulation.

--Average Data packet Reception: It is the number of receptions of the sent data packets in the desired destinations.

--Routing Overhead: It is the number of transfers of network layer controlling packets.

--End to End Delay: A delay which is required for a packet to arrive from the source to the destination.

The compared methods are as follow:

--FMM (Fuzzy Mobility Model).

--OMM( Obstacle Mobility Model ).

--CBMM (Cluster Based-Mobility Model).

--RWMM( Random Waypoint Mobility Model ).

## 4.4 The Results of Simulation of Fuzzy Mobility Model

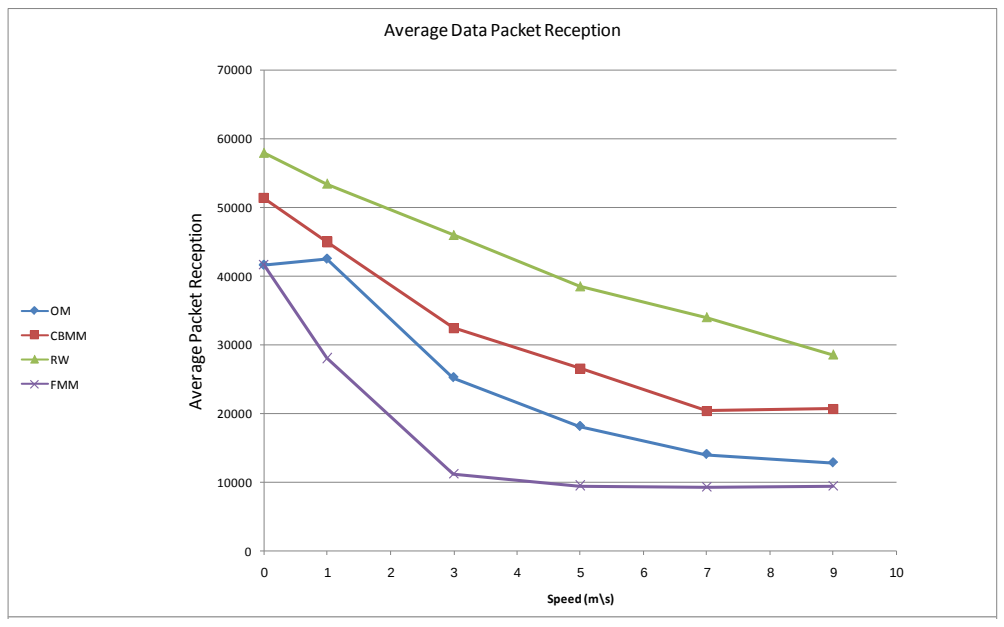

Fig. 5 Average data packet reception.

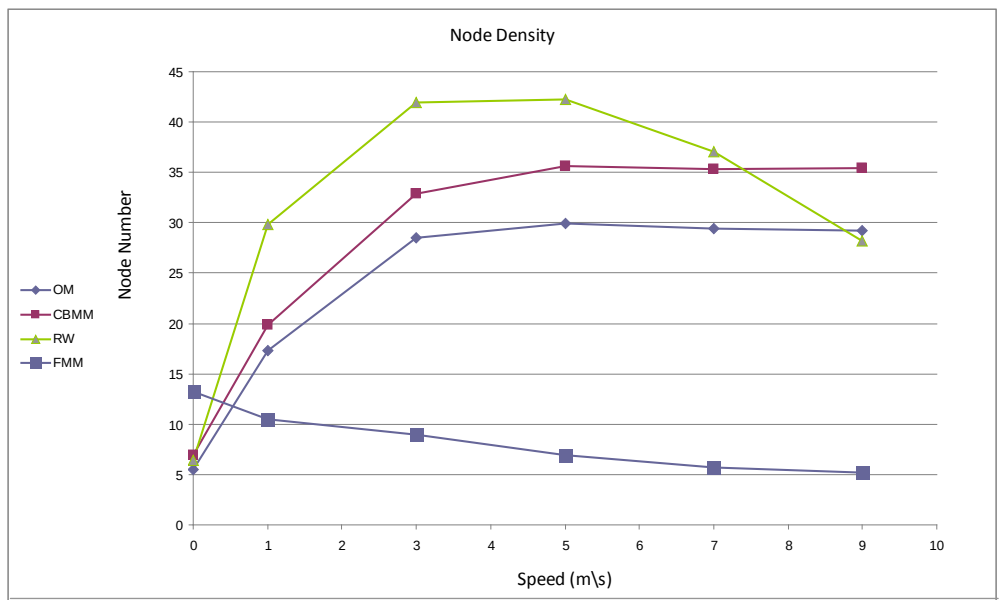

Fig. 6 Node density average.

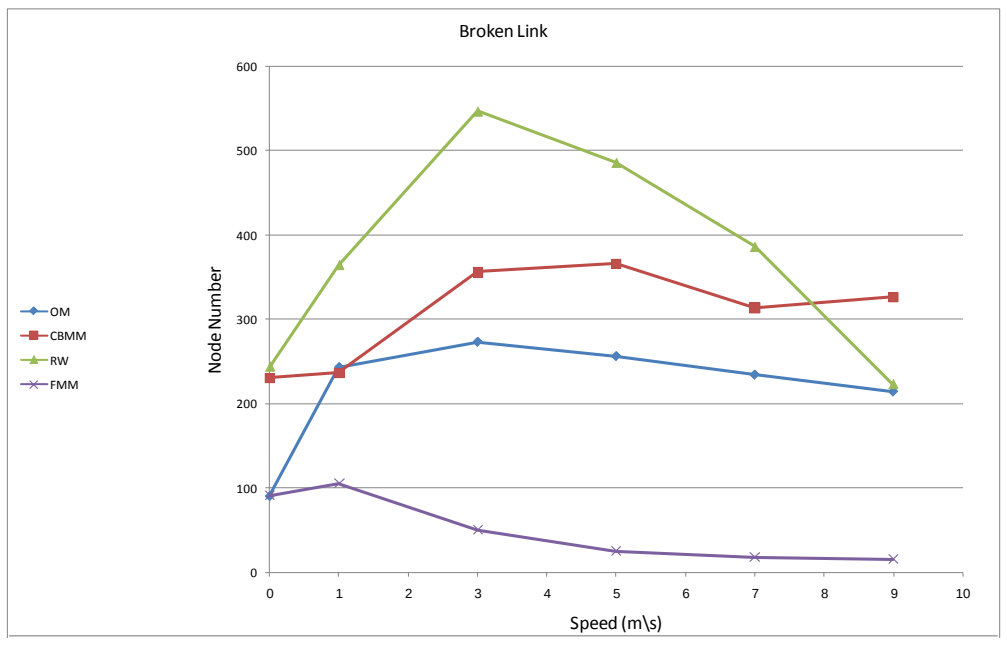

Fig. 7 Broken link average.

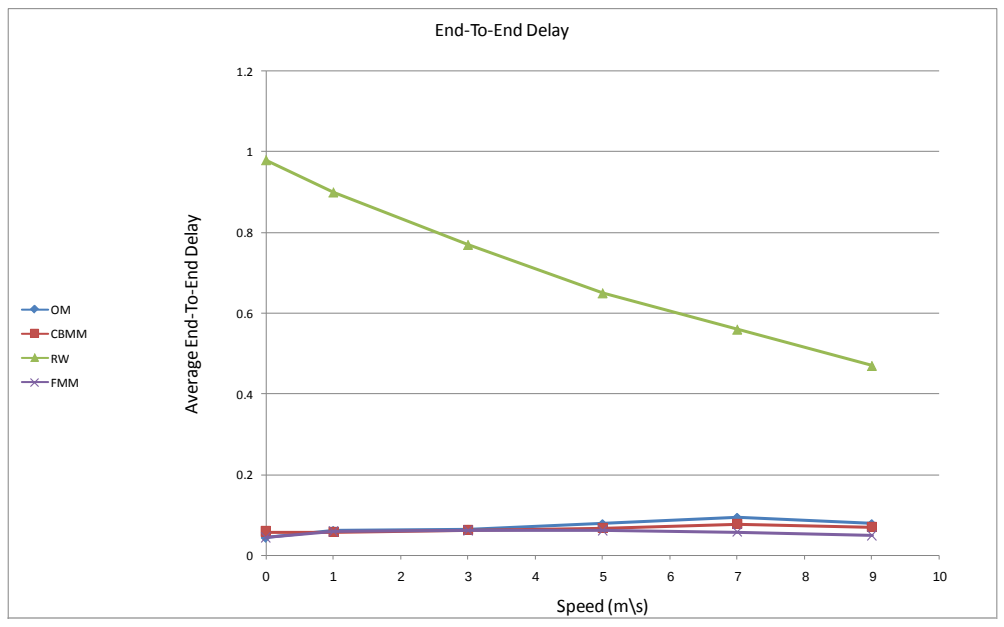

Fig. 8 End to End delay average.

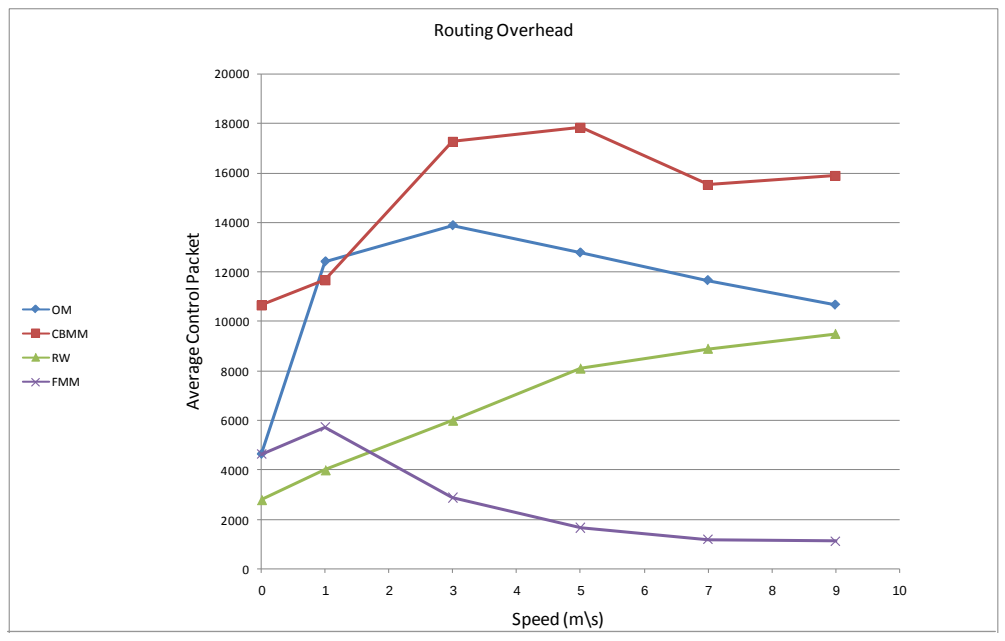

Fig. 9 Routing overhead average.

### 4.4.1 Average Data Packet Reception

Considering figure 5, it can be observed that Average Data Package Reception in RW Mobility Model is better than the other methods. In FMM Mobility Model Average Data Package Reception is lower, because nodes of the same type (for example, Public automobile node) are not beside one another. In CBMM Mobility Model, Average Data Package Reception is better, because nodes of the same type usually are beside one another.

### 4.4.2 Node Density

In RW mobility model node density is better other models, because all parameters (such as destination selection, route selection, and movement speed) are selected randomly. OM mobility model has node density of better rather than FMM mobility model, but the less node density has rather than two mobility models. Figure. 6 shows the average number of each node's neighbors.

4.4.3 Broken Link Average

In FMM, pathways are stable, so fewer number of links break. RWMM has a greater number of broken links and generally OMM and CBMM have a fewer number of broken links than RWMM. It can be concluded that the more is the average number of each node's neighbors in the related mobility model, the more broken links will be. Figure. 7 illustrates broken link average.

4.4.4 End to End Delay Average

In FMM mobility model, has the least end to end delay because pathways are stable. In both models (RW and FMM) there are no obstacles, but whereas the selection of destination and pathway in FMM mobility model against RW mobility model is not random, resulting end-to-end delay of that is lesser. End to end delay in two mobility model CBMM and OM is less than the RW mobility model. End to end delay average is illustrated in figure. 8.

4.4.5 Routing Overhead

From figure. 9, we can conclude that routing overhead in FMM mobility model is the least of other mobility models. But in other models due to pathways stable is less, the mobility models have to be sent more control packets. This makes the routing overhead is more than instead of proposed mobility model.

## 4.5 Conclusion

The focus of this paper is on the pattern of nodes movement in the real environment. The previous mobility models were either unrealistic not including obstacles and pathways, or realistic including obstacles and pathways similar to those of the real environment, but none of them pays any attention to the manner of nodes movement and destination selection.

Considering the examinations have done up to now, each realistic model is composed of 3 sub-models: Environmental sub-model, signal obstruction sub-model and movement pattern sub-model. There is a close relationship among these sub-models. In the proposed mobility model, environmental sub-model and the signal obstruction of obstacle mobility model have been applied, but it has a different movement pattern sub-model.

A fuzzy control system containing fuzzy rules has been used in this mobility model. In this paper, it has been proved that the mobility of a mobile node is fuzzy (imprecise) and also the mobility environment is a fuzzy one. The fuzzy control system used in this paper describes node mobility in an adaptable way to the environment. These rules describe the manner of destination selection. By using a fuzzy control system in the proposed Fuzzy Mobility Model, the movement rules of different types of nodes, depending on the kind of activity and environment and so on, have been imposed. This model also has a knowledge base which is changeable depending on nodes conditions, types of nodes and the environment. Using such knowledge base, movement rules of every environment can be imposed as input on the mobility model in order to consider the movement in that environment.

The type of mobility model, number, type of arrangement, size of obstacles and the speed of nodes movement are the parameters which have a considerable effect on the simulation results.

After simulation, it was found out that not only most of the results in Fuzzy mobility model have improved but also the nearest this model to real world conditions has helped to effectively. This model can help those researchers who would like to implement ad hoc networks protocols.

## References

[1] L. Bajaj, M. Takai, R. Ahuja, K. Tang, R. Bagrodia, and M. Gerla, "Glomosim: A Scalable Network Simulation Environment:, Technical Report CSD, #990027, UCLA, 2003.

[2] The Network Simulator 2, http://www.isi.edu/nsnam/ns.

[3] j. Tian, J. Hahner, C. Becker, I. Stepanov, K. Rothermel, "Graph-based Mobility Model for Mobile Ad Hoc Network Simulation", in the Proceedings of 35th Annual Simulatin Symposium, in cooperation with the IEEE Computer Society and ACM. San Diego, California. 2002.

[4] A. P. Jardosh, E. M. Belding-Royer, K. C. Almeroth, and S. Suri,"Towards Realistic Mobility Models for Mobile Ad Hoc Netwotks", in Proceedings of ACM MOBICOM, San Diego, CA, 2003, pp. 217-229.

[5] M. Berg, M. Kreveld, M. Overmars, O. Schwarzkopf, "Computational Geometry: Algorithms and Applications", Springers Verlog, 2000.

[6] J. Kristoffersson, "Obstacle Constrained Group Mobility Model", in Department of Computer Science and Electrical Engineering Lulea University of Technology, Sweden, 2005.

[7] X. Hong, M. Geral, G. Pei, and C. C. Chaing, "The Performance of Query Control Schemes for the Zone Routing Protocol", in ACM SIGCOMM Describes ZRP Protocol, 1998.

[8] Bittner .Sven, Raffel .Wolf-Ulrich, and Scholz, "Manuel The Area Graph-based Mobility Model and its Impact on Data Dissemination Proceedings" of the 3rd Int'l Conf. on Pervasive Computing and Communications Workshops (PerCom 2005 Workshops) 2005.

[9] Q. Zheng, X. Hong, S. Ray, "Recent Advances in Mobility Modeling for Mobile Ad Hoc Network Research", in ACM-

SE 42 Proceedings of the 42th annual Southeast regional, Huntsville, Alabama, USA, 2004.
[10] Gang Lu, Belis Demetrios, Manson Gordon, “Study on Environment Mobility Models for Mobile Ad Hoc Network: hotspot Mobility Model and Route Mobility Model,” Wireless Com, Hawaii, USA, 2005.
[11] M. Romoozi, H. Babaei, M. Fathi, M. Romoozi, "A Cluster-Based Mobility Model for Intelligent Nodes", in LNCS., Verlag Berlin Heidelberg, 2009, pp. 565-579.
[12] H. Babaei, M. Fathi, M. Romoozi, "Obstacle Mobility Model Based on Activity Area in Ad Hoc Networks", in LNCS., Verlag Berlin Heidelberg, 2007, pp. 804-817.
[13] F. Bai, N. Sadagopan, A. Helmy, "The Important Framework For Analyzing The Impact of Mobility on Performance of Routing Protocols for Ad Hoc Networks", in Proceedings of IEEE INFOCOM, San Francisco, CA, 2003, pp. 825-832.
[14] S. Marinoni, H. Kari, "Ad Hoc Routing Protocol Performance in a Realistic Environment", in Proceeding of the 5th IEEE International Conference on Networking (ICN) , Mauritius, 2006.
[15] Wang, Lie-Xin, “A course in fuzzy systems and control.”

**Alireza Amirshahi** is currently Ms student at Islamic Azad University (Arak branch) in Iran. He was born in Kashan at 11 September 1973. He received a Bs in software engineering from the Islamic Azad University (Kashan branch) at 2002. He has taught in the areas of computer architecture and logic circuits and his research interests including computer architecture and Ad hoc networks.

**Mahmood Fathy** received his BSc in electronics from Iran University of Science and Technology in 1985, MSc in computer architecture in 1987 from Bradford University,
United Kingdom and PhD in image processing computer architecture in 1991 from UMIST, United Kingdom. Since 1991, he has been an associate professor in the Computer
Engineering School of IUST. His research interests include image and video processing, computer networks, including wireless and vehicular ad hoc network and video and image transmission over the Internet.